\pdfoutput=1 

\documentclass[a4paper,11pt,article]{memoir}
\usepackage{hyperref}
\hypersetup{pdftex,pdfborder={0 0 0}}

\usepackage{ecis2020}

\usepackage{makecell}
\usepackage{cleveref}
\usepackage{booktabs}
\usepackage{siunitx}
\usepackage{rotating}
\usepackage{tabularx}

\maintitle{A network-based transfer learning approach to improve sales forecasting of new products}
\shorttitle{Transfer learning for sales forecasting} 
\category{Complete Research} 

\authors{
Karb, Tristan, Karlsruhe Institute of Technology, Germany, tristan.karb@alumni.kit.edu 

Kühl, Niklas, Karlsruhe Institute of Technology, Germany,  niklas.kuehl@kit.edu 

Hirt, Robin, Karlsruhe Institute of Technology, Germany, robin.hirt@kit.edu 

Glivici-Cotruță, Varvara, Schwarz IT KG, Germany, varvara.glivici-cotruta@mail.schwarz
}

\shortauthors{Karb et al.} 

\begin{document}

\begin{abstract}\noindent
Data-driven methods---such as machine learning and time series forecasting---are widely used for sales forecasting in the food retail domain. However, for newly introduced products insufficient training data is available to train accurate models. In this case, human expert systems are implemented to improve prediction performance. Human experts rely on their implicit and explicit domain knowledge and transfer knowledge about historical sales of similar products to forecast new product sales. By applying the concept of Transfer Learning, we propose an analytical approach to transfer knowledge between listed stock products and new products. A network-based Transfer Learning approach for deep neural networks is designed to investigate the efficiency of Transfer Learning in the domain of food sales forecasting. Furthermore, we examine how knowledge can be shared across different products and how to identify the products most suitable for transfer. To test the proposed approach, we conduct a comprehensive case study for a newly introduced product, based on data of an Austrian food retailing company. The experimental results show, that the prediction accuracy of deep neural networks for food sales forecasting can be effectively increased using the proposed approach.
\end{abstract}


\chapter{Introduction}

Knowledge about future development of demand and sales is a critical success factor for businesses across all domains, e.g. manufacturing, wholesale or retail of products. This knowledge is essential to optimize production, logistics and stock management with the objective to maximize sales and minimize costs along the product value chain. Due to the perishability of the goods and the associated short shelf-life, correct forecasts are of particular importance in the food industry \citep{Doganis.2006}. In the U.S., food waste accounted for ten percent of all goods produced in food retailing in 2010 \citep{JeanC.Buzby.}. These figures outline the business value and relevance of improved sales forecasts in food retail, both from a business and sustainability perspective. Sales forecasting can be based on expert knowledge, statistical time series forecasting techniques or machine learning (ML) models \citep{Chu.2003}. The knowledge of skilled sales managers is a valuable asset in this domain and expert-based systems can achieve good prediction performances. However, in order to be less dependent on the knowledge and availability of these experts, data-based prediction models have increasingly been developed and implemented. While originally time series models were used primarily, there is now a clear trend towards the implementation of ML models in the field of food sales forecasting \citep{Tsoumakas.2019}. This trend is based on the generally good performance of ML models in this domain, as well as the increased availability of internal and external data and suitable ML tools \citep{Pedregosa.2011}.
However, the performance of ML models strongly depends on the quality and availability of the data \citep{Witten.2016}. With newly introduced products, the amount of available data is limited and often special events---such as promotions or holidays---with a strong influence on daily sales have not been observed in the available data \citep{Kahn.2014}. Under these conditions, stand-alone statistical and ML models tend to have poor prediction quality, making expert knowledge a valuable asset to improve forecasting quality. This expert knowledge can be used for selecting similar reference products as proxy to predict sales of new products in general or for adjusting forecasts during special events. These actions are based on implicit and explicit knowledge and rely on judgment and assumptions made by the expert \citep{Kahn.2014}. It must therefore be considered that human bias may be present and that decisions are based on limited information. Simultaneously, the business success is strongly dependent on the permanent availability of these experts. The concept of Transfer Learning (TL) provides an analytical approach for ML models to mimic the human behavior of transferring knowledge in-between similar problem domains. The general idea is to improve a model in a \textit{target} domain (in our case sales data of a new product) by transferring information from a related \textit{source} domain (sales data of similar products) \citep{Weiss.2016}. In recent years, there has been considerable interest in the field of TL across various forecasting domains, such as energy demand \citep{Ribeiro.2018}, energy production \citep{Hu.2016} and stock prices \citep{Li.2018}. The potential of transferring machine learning models is also motivating researchers and practitioners to use neural networks---which are considered to be black boxes---as information carriers in business networks, enabling an inter-organizational learning \citep{HirtKuehl2018_1000086441}. These promising insights motivate our proposal to apply a TL approach for the problem of forecasting new product sales in the food retail domain. Transfer learning could improve the forecast of sales in this domain in a significant manner and, thus, increase economic efficiency tremendously. We are interested in the effect of such a transfer in comparison to machine learning models, which are not leveraging transfer learning. Another field of interest is understanding when it makes sense to transfer. Where previous work is already examining indicators based on data and even neural network structures \citep{hirt2020sequential}, we want to further examine indicators which are based on characteristics of the real-life representation of the data at hand. 
Therefore, our work contributes to the body of knowledge in three ways. First, we propose and evaluate a network-based TL approach for multivariate sales forecasting of newly introduced products, instantiated in the food retail area. 
Second, we expect the performance of transferred models to depend on the similarity between the two domains of source and target. We hypothesize different ways to capture product similarity, and, most importantly, to identify which products are most suitable for the transfer. Third, we illustrate the approach within a comprehensive field study in cooperation with a food retailer in Austria, which aims to investigate the sales forecasting of a newly introduced product. We evaluate different TL strategies in the presence of insufficient data and special events. The experimental results show that the prediction errors are significantly reduced using the proposed technique.

The remainder of the paper is structured as follows. In Section \ref{2}, foundations of sales forecasting and TL are described. In Section \ref{3}, we outline the Design Science Research-based research design as well as our technical methodology of our proposed approach, which was applied in the case study in Section \ref{4}. The results of the case study are discussed in Section \ref{5}. Finally, we offer some concluding remarks regarding this research in Section \ref{6}.
\chapter{Foundations}\label{2}
\section{Food retail sales forecasting}
Sales and demand forecasting methods can be differentiated by the underlying business objective, the models used for prediction and the included data. The business objective determines the temporal granularity of the forecasting model \citep{Tsoumakas.2019}. Whereas monthly or quarterly predictions are used to make strategic business decisions \citep{Doganis.2006}, weekly predictions are used for stock management \citep{Zliobaite.2012}. Sales of products, which are produced daily (e.g. bakery products) or have a short shelf-live (e.g. vegetables, fruits), are predicted on a daily or hourly basis \citep{Chen.2010}. 

 Traditional statistical time series forecasting models are widely considered in many real-world applications for sales forecasting. Commonly used models are moving average (MA) \citep{Zliobaite.2012}, auto-regressive moving average (ARMA) \citep{Choi.2014} and auto-regressive integrated moving average (ARIMA) \citep{Bakker.2009}. Compared with more complex approaches, advantages of time series forecasting methods depict better interpretability and an easier implementation \citep{Arunraj.2016}. However, these models are restricted due to their linear structure \citep{Choi.2014} and are limited in their ability to include external variables \citep{Arunraj.2016}. Demand in food retailing is influenced by a variety of internal (prices, promotions, substitutes, etc.) and external factors (weather, holidays, competitors, etc.) \citep{Vorst}. While ML models are designed to process complex multi-variate data, both human experts and statistical forecasting methods are limited in their ability to integrate all available data in its full complexity into their forecasts. This distinguishing characteristic as well as the generally increased availability of internal and external data make ML models increasingly important in the field of sales forecasting \citep{Tsoumakas.2019}. Recent studies on ML for demand and sales forecasting indicate that ML methods outperform traditional time series forecasting methods \citep{JuanPabloUsugaCadavid., Pavlyshenko.2019}. Popular algorithms for retail sales forecasting are decision trees \citep{Castillo.2017} , logistic regression \citep{Bakker.2009}, ensemble models \citep{Meulstee} and deep neural networks (DNNs) \citep{Slimani.2017, Lu.2016, Liu.2017}.

\section{Transfer Learning}
For humans, TL is an automatic mechanism when it comes to solving new problems. Humans automatically apply previously learned knowledge to solve new problems better or faster \citep{Pan.2010}, which is illustrated by the following example of two people learning a new instrument. A person who can already play the guitar will learn to play the violin much faster than a person without any experience with instruments. Consciously and unconsciously, the experienced person transfers knowledge from a \textit{source domain} \textit{D\textsubscript{s}} and a \textit{source task} \textit{T\textsubscript{s}} (in this case playing the cello) to the \textit{target domain} \textit{D\textsubscript{t}} and for the \textit{target task} \textit{T\textsubscript{t}} (playing the violin). In the context of ML, TL is a promising approach when either sufficient (labeled) training data is missing in the target domain or when training a dedicated model in the is computationally expensive \citep{Weiss.2016}. 

The three main research areas in the context of TL are \textit{what} to transfer, \textit{how} to transfer and \textit{when} to transfer \citep{Pan.2010}. For the work at hand, the most relevant are the latter two. \textit{``How to transfer''} corresponds to the development of algorithms and approaches to transfer knowledge. In \cite{Weiss.2016}, different TL approaches are divided into four general categories. In \textit{instance-based} TL, instances from the source domain are transferred to the target domain, re-weighted in order to correct marginal distribution differences and directly used for training. \textit{Feature-based} TL is used to map the features of \textit{D\textsubscript{s}} and \textit{D\textsubscript{t}} into a common latent feature space (symmetric transformation) or to re-weight the features of \textit{D\textsubscript{s}} to match \textit{D\textsubscript{t}} (asymmetric transformation). \textit{Parameter-based} TL describes approaches to transfer parameters of models trained in source domains to models trained in the target domain. \textit{Network-based} TL is a subcategory of parameter-based TL, where the network parameters of a DNN, including its structure and connection parameters, trained in \textit{D\textsubscript{s}} for T\textsubscript{s} are transferred to be used in \textit{D\textsubscript{t}} \citep{Tan}. \textit{Relational-based }TL is the fourth category. In this category knowledge is transferred based on relational patterns between source and target domain. \textit{``When to transfer''} refers to the fundamental question of when the knowledge transfer can be applied effectively. The TL concept is based on the assumption that there are similarities between \textit{D\textsubscript{s}} and \textit{D\textsubscript{t}} and that it can be useful to transfer knowledge between these domains. At the same time not all available source domains are suitable for the transfer and in the worst case the performance of T\textsubscript{t} is degraded by the transfer. Thus, it is essential to prevent this negative transfer and to identify the source domains that are best suited for the transfer. In this regard, research is aiming towards answering that question from a data perspective by identifying relations between data similarity and the effect of a transfer. Different proxies, such as data projections in form of a maximum mean discrepancy or even neural networks are used to indicate the utility of a DNN transfer \citep{hirt2020sequential}. In contrast, this work focuses on examining the influences of real-world representational characteristics of data to a network-based transfer in a use case ex post. Insights could then lead to a better understanding of the transfer and a possible application of ex ante methods. 

TL has been successfully applied in many real-world applications. Popular domains are image and video processing \citep{shao2014transfer}, natural language processing \citep{lu2015transfer} or recommender systems \citep{fernandez2012cross}. 
In the domain of (time series) forecasting, TL has been applied in the context of energy demand \citep{Zeng.2019} and power output \citep{Cai.2019} prediction, weather forecasting \citep{Huang.} and the prediction of oil \citep{Xiao.2017} and stock \citep{Li.2018} prices. Instance-based and parameter-based approaches were primarily implemented in the case studies. In \citeauthor{Afrin.2018} (\citeyear{Afrin.2018}) an instance-based TL approach was implemented to improve the predictions of new product sales in the automotive industry. \cite{NikolayLaptev.} have shown, that network-based TL with DNNs is suitable for time series forecasting and that TL models are capable of outperforming regularly trained DNNs in multiple domains. These results coincide with other case studies related to forecasting in which a network-based TL approach with DNNs was used \citep{Qureshi.2017, Hu.2016, Liang.2018, MarcusVos.2018}.

Previous work is also focussing to utilize TL as a means for information exchange in business networks to overcome data confidentiality challenges \citep{HirtKuehl2018_1000086441}. Abstract neural nets can be exchanged between organizations instead of the raw data to gain superior prediction models. By sequentially transferring models across multiple data sets and organizations, network-wide learning can be achieved without the exposure of raw data \citep{Hirt2019}. 

In contrast, this work focuses on determining influence factors for TL based on real-world characteristics of the machine learning problem and examining possible relationships. Furthermore, this paper is the first to develop a network-based TL approach for new product sales forecasting in the food retail domain. 

\chapter{Research Design and Methodology}\label{3}

As an overall research design, we choose Design Science Research (DSR) with a strong focus on the (technical) evaluation of the developed artifact. According to \cite{Hevner.2010}, a DSR project should cover at least three cycles of investigation, a rigor cycle (which we covered in the previous section), a relevance cycle (motivated in the case study in Section 4) and one or multiple design cycles (which is covered in this section). The research questions, which are derived from theory and practice and which we aim to answer in the design phases can be formulated as follows:
\begin{itemize}
    \item \textbf{RQ1:} How well do transferred models with different degrees of freedom (TL\_models) perform in comparison to models without transfer (models w/o\_TL) for a newly introduced product?
    \item \textbf{RQ2:} Which conclusions can we draw in regard to the similarity of source and target domain after experimenting with different transfer options?
    \item \textbf{RQ3:} Which characteristics of product groups could give insights on the success of transferring models in advance?
\end{itemize}

According to \cite{GregorHevner}, the developed artifact within DSR can be considered as knowledge contribution, if the proposed artifact either provides a new solution for a known problem (improvement), a new solution for new problems (invention) or extend known solutions to new problems (exaptation). In our case we want to develop a new solution for the known problem of forecasting new product sales, as described in \cite{Kahn.2014}. Our proposed artifact of network-based TL is best described as a \textit{method}, since it consists of ``actionable instructions that are conceptual'' \citep{Peffers2012}. It is designed to investigate if and how prediction accuracy of DNNs for sales forecasting of newly introduced products can be improved with network-based TL.

To allow for a common understanding, we elaborate on the decisions made within our design project in the upcoming paragraph. DNNs were chosen as forecasting models for two reasons. They are the most popular ML model in food sales forecasting \autocite{Zliobaite.2012} and the described network-based TL approach is specially designed for neural networks. 
Recent studies in food sales forecasting propose various different neural network architectures. Multi-layer perceptrons (MLP) with back-propagation were implemented to predict monthly aggregated retail sales \autocite{Chu.2003}, weekly sales of new products \autocite{Livieris.2019} or daily sales \autocite{Slimani.2017}. The implemented MLP in \citeauthor{Chen.2010} (\citeyear{Chen.2010}) outperformed logistic regression and moving average models in predicting daily sales of fresh foods in Taiwanese supermarkets. More complex DNN architectures were implemented in \citeauthor{Liu.2017} (\citeyear{Liu.2017}), namely Auto Encoders (AE) and Long Short Term Memory Networks (LSTM). They used sales and meteorological data to predict daily sales in Japanese supermarkets. 
For our approach a MLP with back-propagation as network architecture was selected. A MLP is a feed-forward neural network. It consists of one input layer, one or more hidden-layers and one output layer. The layers consist of interconnected neurons, which are connected by weights. The output of each node is computed by the sum of its inputs, modified by a linear or nonlinear activation function. This output is fed forward to the next layer of the network. The back-propagation algorithm is used to train the network. The algorithm propagates the error back through the network and adjusts the weights to minimize the overall error \autocite{M.WGardner.1998}. 

\citeauthor{Tan} (\citeyear{Tan}) and \citeauthor{JasonYosinski.} (\citeyear{JasonYosinski.}) provided comprehensive studies on recent implementations of network-based TL. As described in Section \ref{2}, TL has been widely implemented in image recognition, video and signal processing. Most popular within these domains is the implementation of network-based TL with convolutional neural networks (CNN). In other studies, the network-based transfer was conducted with different DNN architectures, such as AEs or LSTMs. 
What makes neural networks suitable for the application of TL is their ability to learn general features on the first layers of the network and task specific features on the last layers \autocite{JasonYosinski.}. In this way, DNNs trained for image processing, regardless of the specific task, learn to abstract forms and color structures on the first layers \autocite{JasonYosinski.}. Only on the last layers the categorization into certain classes is learned \autocite{shin2016deep}. The same effect can be observed when processing time series data with CNNs, where individual snapshots of the time series are processed. Again, the CNNs learn to abstract generic features, such as the occurrence of sequences and seasonalities. In this domain pre-trained CNNs have better generalization performances than CNNs trained without TL, even if the transfer is conducted with time series from dissimilar domains \autocite{Fawaz.2018}. As \cite{JasonYosinski.} show, DNNs with transferred features perform better than those with randomly initialized weights, even after substantial fine-tuning on the target task. They have also discovered that the effect of TL diminishes the more dissimilar the source and target tasks are. Besides transferring the first layers containing general features, it is worthwhile to transfer the task-specific features, as long as there is a high similarity between the tasks. For instance, when using CNNs with TL in image processing, often all layers of a pre-trained network are transferred. The weights of the first layers are often completely frozen during the training and the only the last layers are fine-tuned for the respective task. Consequently, the main objectives in the implementation of network-based TL are to find tasks that are as similar as possible and to determine on which layers the network is fine tuned during training. The objective of our research is to investigate if these characteristics of DNNs are reproducible in the domain of sales forecasting with multivariate data. For this purpose, we examine the prediction results of different settings for fine tuning and training on the transferred layers.

In other studies applying TL approaches in the forecasting domain, similar wind farms have been used as source domain predict the power output of new wind farms with few training data \autocite{Cao.}, similar cities have been used as source domain to improve holiday load forecasting for cities with insufficient data \autocite{Zeng.2019}. In our case, we want to investigate if this concept is applicable with retail products. In related work, three different approaches are described to deduce the transferability from the similarity of source and target, namely data similarity, task similarity and model similarity \citep{hirt2020sequential}. In our case we focus on data similarity. Subsequently we want to investigate, if transferability can be derived from a priori known information of the source and target products. We want to investigate if the prediction performance of models with TL is related to similarities of product attributes, such as article family, mean hourly sales or price. For this, objective product similarities and prediction performances of different source products are put into context and analyzed, whether it is possible to infer from product similarities to task similarities. 

\chapter{Case study}\label{4}
The application domain of our DSR project is situated in the sales forecasting system for bakery goods of a large food retailing company in Austria, which provided the data. The already implemented ML-based forecasting system yields good performances in predicting in-stock products. As part of our relevance cycle we conducted interviews with experts in this domain. In this cycle, problems related to forecasting new product sales were described and the potential to improve the prediction performance of new products was identified. 

The target domain of the TL approach is a bakery product that was newly introduced in 99 stores. The data of the first 17 weeks after the introduction were available for our case study. The source domains are 14 products that are listed in the same subgroup of bakery products as the target product and have been sold continuously for at least two years. Since the source data is available over a long period of time, the effects of promotions, holidays and seasonalities are represented in this data.
Bakery goods have a short shelf live and are freshly baked several times a day in each store. Therefore, the objective and output variable of the forecasting model is to predict the hourly sales of the product in each store. These forecasts are made on Sundays for the entire following week.
In order to access more training data and guarantee a better generalization of the model, only one holistic model is trained for all stores. To allow for a training period, the sales starting from week 3 are to be predicted. The first week is needed for the calculation of lag and aggregation features. Therefore, the model can be trained for the first time with the data of the second week. 
\begin{figure}[htbp]
 \centering
 \includegraphics[height=3cm]{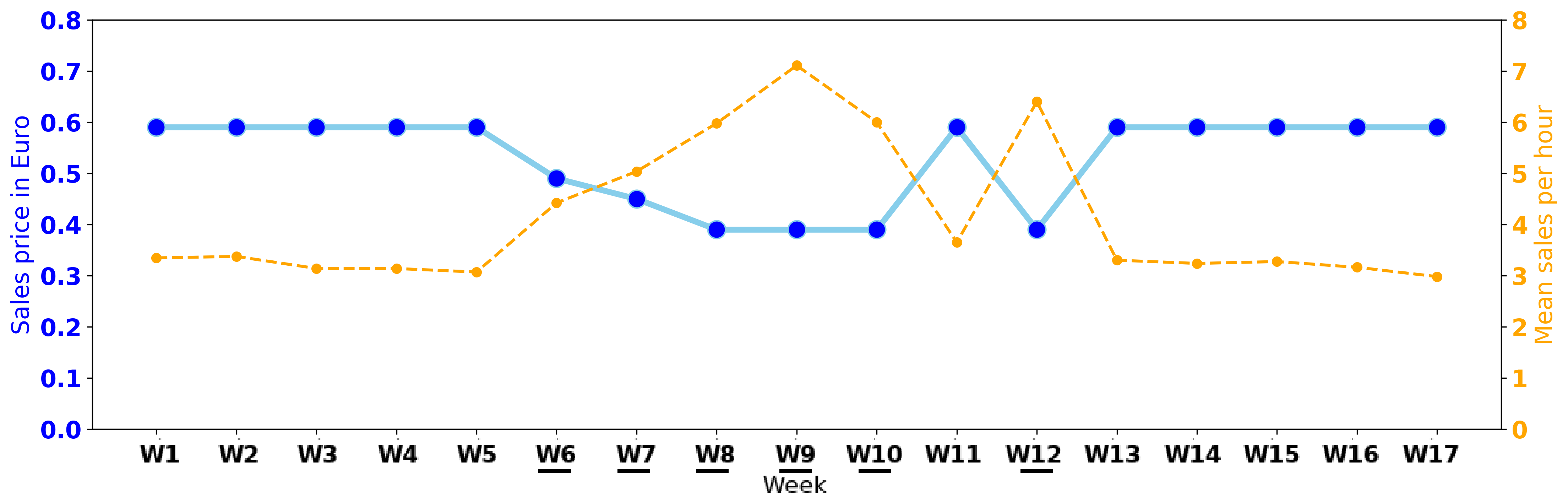}
 \caption{History of sales price and weekly average of sales per hour in the period examined. Weeks with promotions are underlined} 
 \label{fig_sales}
\end{figure}
In regards to meta data of the product, it was in promotion for several weeks in which it was offered with a lower sales price. The product has on average significantly more sales per hour in these weeks than in the weeks without promotions (see Figure \ref{fig_sales}). The low amount of training data in general and the occurrence of unseen promotions with a significant influence on the target variable represent the challenges for the forecasting models in this case study. 

In the remainder of this section, the data and features used will be analyzed in more detail. Next, the implemented models are described and the metrics of the evaluation are defined. At the end of this section, the results of the conducted case study are presented.

\section{Descriptive Data Analysis and Similarity Clustering}

 The hourly sales data is aggregated during the opening hours of every store during the examined period of 17 weeks. A total of 26 features were used as input data for the prediction models. These features can be divided into six different types: date, price/promotion, identity, lag, aggregate and external features. A more detailed description for all features is depicted in Table \ref{tab:features}. 
 \begin{table}[htbp]
\centering
\begin{tabular}{ccc}
  \hline
            Category & Features & Description \\ 
  \hline

Date & 8 & Derived date features\\
Price/promotion & 6 & Sales price, promotional status and derived features \\
Identity & 2 & Store number and state of location \\
Lag Features & 4 & Moving averages over different time windows and aggregation levels\\
Aggregates & 3 & Averages over all available data on different aggregation levels\\
External Features & 3 & Days until/since public holiday, school holiday \\ 
  \hline
\end{tabular}
\caption{Description of input features} 
\label{tab:features}
\end{table}
 The lag and price/promotion features are of particular importance for the case study and are therefore explained more in detail. The utilization of aggregated lag variables are a promising strategy for sales forecasting \autocite{Doganis.2006, Meulstee,Zliobaite2009}. Retail sales follow certain hourly, daily and weekly patterns. Thus, sales on the same day of the previous week are on average better indicators than sales on the previous day \autocite{Tsoumakas.2019}. This is particularly evident in the example of sales on Fridays and Saturdays. In our case, the four lag features are moving averages over different time windows, aggregating the average sales on hourly, daily or weekly levels for each store.
 The price/promotion features include the sales price, the information whether the product was under promotion and the respective promotion price. As visualized in Figure \ref{fig_sales}, the product was under promotion during weeks 6,7,8,9,10 and 12, in which the average hourly sales are remarkably higher. To take account of this general difference in average hourly sales when calculating the lag features, only weeks with the same promotional status as the respective week are included in the aggregation. For example, weeks 3 and 4 are used to calculate the two-week moving average values for week 5 (without promotion). For week 12 (with promotion), the values of the last two weeks with promotion are used and thus weeks 9 and 10.

As described in Section \ref{3}, the success of the transfer is highly depended on the similarity between source and target. We analyze three different dimensions to compare product similarities, based on the information available at the time of the first forecast in week three. 
For the target product, additionally to the sales price, the mean and standard deviation of the hourly sales as well as the share of promotion in week one and two, are known at this point (Table \ref{tab:product_data}). The values for the source products in this table are based on two years of available data. 
\begin{table}[htbp]
\centering
\sisetup{separate-uncertainty}
\begin{tabular}
  {@{} 
    l
    c
    c
    c
    c
    c
    c
  @{}}
  \toprule
    & \multicolumn{3}{c}{Values} & \multicolumn{3}{c}{Cluster} \\
    &  \multicolumn{1}{c}{Mean hourly sales} & {Sales price} & {\% Promotion data} & {Mean sales} & {Price} & {Promotion}\\

  \midrule
Target & \num{3.13+-2.44} & 0.59 & 0.00 & Low & Medium & Low\\   \bottomrule
Source 1 & \num{2.42+-1.88} & 0.99 & 0.01 & Low & High & Low\\
Source 2 & \num{3.8+-3.47} & 0.79 & 0.02 & Low & High & Low\\
Source 3 & \num{4.24+-3.28} & 0.59 & 0.04 & Low & Medium & Medium\\
Source 4 & \num{4.66+-3.67} & 0.55 & 0.02 & Low & Medium & Low\\
Source 5 & \num{5.32+-4.27} & 0.59 & 0.09 & Medium & Medium & High\\
Source 6 & \num{5.7+-4.56} & 0.55 & 0.04 & Medium & Medium & Medium\\
Source 7 & \num{5.77+-4.79} & 0.39 & 0.05 & Medium & Low & Medium\\
Source 8 & \num{5.77+-4.61} & 0.49 & 0.09 & Medium & Low & High\\
Source 9 & \num{5.92+-4.80} & 0.45 & 0.04 & Medium & Low & Medium\\
Source 10 & \num{6.46+-5.20} & 0.49 & 0.06 & High & Low & Medium\\
Source 11 & \num{7.01+-5.59} & 0.39 & 0.06 & High & Low & Medium\\
Source 12 & \num{7.06+-5.57} & 0.69 & 0.05 & High & High & Medium\\
Source 13 & \num{7.37+-7.28} & 0.49 & 0.06 & High & Low & Medium\\
Source 14 & \num{7.46+-5.62} & 0.59 & 0.05 & High & Medium & Medium\\
\end{tabular}
\caption{Product specific data and assigned clusters} 
\label{tab:product_data}
\end{table}
Within the three dimensions, the source and target products are assigned to clusters based on the respective values. This clustering approach is used to systematically search for the most suitable source models. The objective is to contextualize the similarity of source and target products in different dimensions with the prediction performance of the TL\_models. In this way, we want to test whether suitable models can be identified with the apriori known information. The target is listed for 0.59 Euro, five source products with a very similar price are assigned to the same cluster. Interestingly, the target has the lowest mean hourly sales in this cluster. source 1 and source 2 have the highest sales prices. The mean hourly sales of source 10, source 11, source 12 and source 13 show that the price is nevertheless not decisive for the level of sales. Furthermore, the source products differ in the proportion of promotion data. While in source 5 and source 8 nine percent of the data is promotion data, in source 1 it is only one percent. The product data, similarities and assigned clusters will be further discussed in Section \ref{results}.

\section{Predictive Model}
To identify a network architecture that delivers good results in all source domains, we applied cross-validation for different network configurations and compared the performances across all source domains. It must be emphasized that the aim of the case study was not to find the optimal network architecture for the prediction problem, but to investigate the proposed TL approach. Therefore, it has to be assumed that the implemented architecture is not optimal, but provides sufficiently good results for our experiment.

\begin{figure}[htbp]
 \centering
 \resizebox{\textwidth}{!}{
 \includegraphics{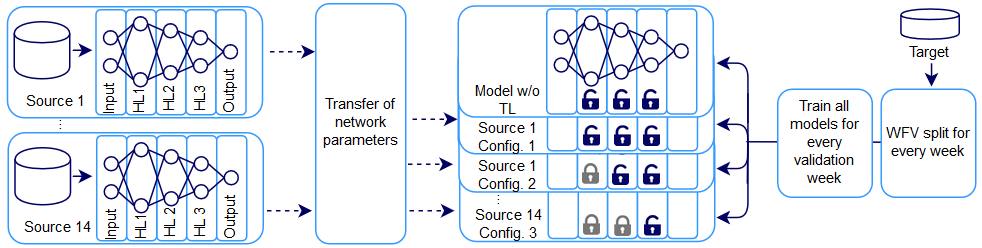}}
 \caption{Process for training source models, transferring the parameters and training the target models} 
 \label{fig_TL_pipeline}
\end{figure}

The implemented MLP has one input layer with 26 nodes, corresponding to the 26 input features, three hidden layers and one output layer with one node. The number of nodes in the hidden layers are 256, 128 and 64. As activation functions on the hidden and output layers tanh, ReLu, tanh and tanh were used. We prevented overfitting dropouts during training according to \autocite{Dropout} with a parameter value of 0.2 as well as an early stopping method \autocite{Prechelt.1998}. Both the source and target products were trained on this network architecture and optimized with the ADAM optimizer \autocite{Kingma.12222014}. For each of the 14 source products one model is trained independently and the network architectures, including all parameters, were saved after training (Figure \ref{fig_TL_pipeline}). At this point the knowledge transfer is conducted by transferring these parameters.
Subsequently the networks were trained with the target data. For training the TL\_models, we defined three different configurations. In config 1, all layers of the network are trainable. In this configuration, the transfer can be seen as a pure weight initialization of the network. In config 2 the first layer is frozen and the network is trained on the other two layers. In config 3. the first two layers are frozen and the network is trained on the last layers. In this configuration, the network has the fewest degrees of freedom to train on the target data. For training the models a walk-forward validation (WFV) was used. The WFV is used for time series data to ensure that training is only done with data that already exists at the time of the prediction \autocite{stein2002benchmarking}. For example, for the prediction of week 8 all data of weeks 2 to 7 are used in our case. For every step of the WFV the model w/o\_TL is trained and the 14 TL\_models are trained individually, every model once in all three described configurations. Subsequently every individual model is used to predict the respective validation week of the WFV. To evaluate the model performance during training, the data was randomly split in 70\% training and 30\% testing data in every step of the WFV. To validate the prediction results, 30 bootstrap random splits were applied and the models trained on each of these splits. 

To ensure the validity of the ML models as predictor, a two week moving average, aggregated on daily level, is used as naive baseline, which is a common baseline in food sales forecasting \autocite{Bakker.2009}. We chose the mean squared error (MSE) as evaluation metric, which is computed as follows:
\begin{equation}
  MSE = \frac{1}{n}\sum_{i=1}^{n}(y_i - \hat{y_i})^2
\end{equation}
where n is the size of the sample, y\textsubscript{i} is the i-th observed value and \^y\textsubscript{i} is the i-th predicted value. MSE is widely used as metric in time series forecasting and food sales prediction and suitable, if only one product is to be predicted \citep{Bakker.2009}.

\section{Results}
\label{results}

\begin{figure}[htbp]
 \centering
 \resizebox{\textwidth}{!}{
 \includegraphics{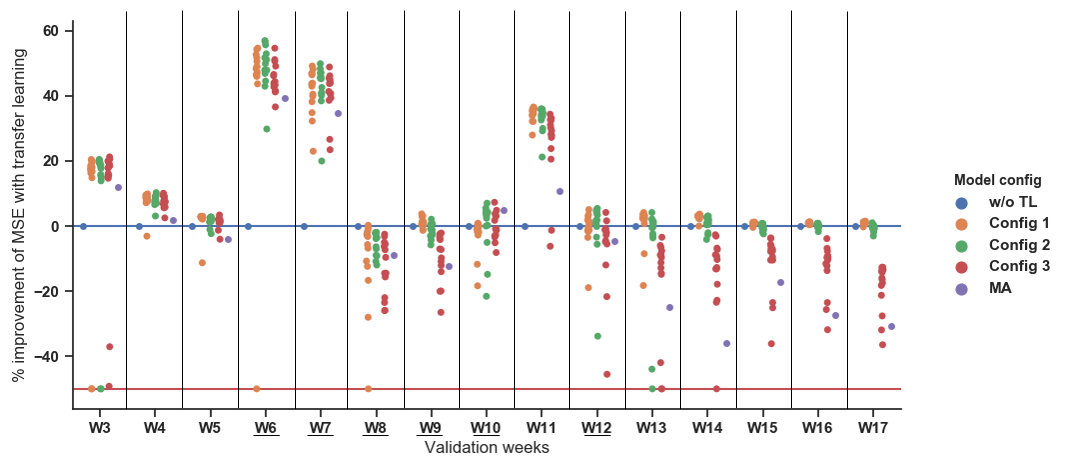}}
 \caption{Comparison of the average performances of the model w/o\_TL and the 14 source models in each of the three transfer configurations, after training on 30 random splits. The performance of the model w/o\_TL is used as baseline and the percentage improvement of the TL\_models is shown} \label{fig_swarm_total}
\end{figure}

The DSR paradigm emphasizes the importance of a comprehensive evaluation \citep{Venable2016}. We discuss this evaluation in this subsection as well as in the upcoming discussion (\Cref{5}). The results of the conducted experiments are summarized in Figure \ref{fig_swarm_total}: the MSE of the model w/o\_TL is utilized as baseline and the percentage difference of all TL\_models and the MA are visualized. Four observations can be derived from Figure \ref{fig_swarm_total}. 
First, we can demonstrate the feasibility to improve the prediction performance with TL in many cases. Especially in the first weeks in general and the first weeks with promotion significant performance improvements can be observed. For both averages over the weeks with promotion and over the whole period, all TL\_models except one are better than the model w/o\_TL. In the weeks without promotion the differences are less clear and a total of eight models have a worse performance on average. 
When analyzing the performances in the first two weeks (W3 and W4) and the first two weeks with promotions (W6 and W7), it should be noted that the model w/o\_TL performs significantly worse than the naive baseline. Most TL\_models outperform the MA baseline, whereas the model w/o\_TL has a worse performance in the first two weeks and the first two weeks with promotions. Most TL\_models already exceed the MA baseline in these phases and a performance advantage can also be observed, but less evident than Fig. \ref{fig_swarm_total} suggests. This means that the advantage of the TL\_models is partly due to the poor handling of the model w/o\_TL with previously unseen values of the input variables. 
Second, the degree of improvement decreases with increasing duration and thus increasing data availability in the target domain. 
The improvement in W5 is already considerably lower, but will increase again in the following weeks due to the promotion. Interestingly, in W8 only one TL\_model is better than the model w/o\_TL, so the transfer worsens the performance this week. Furthermore, it can be seen that the second time a promotion phase occurs in W12, the advantage of the TL\_models is also significantly lower than during the first promotion phase. 
Nevertheless, even in the last validation weeks improvements with the TL\_models are possible for some models and configurations. 
This leads to the third observation: there is a clear difference in the performance of the different configurations. Figure \ref{fig_swarm_total} shows a significantly worse performance of the models trained with config 3, especially during the later weeks. We can conclude the low degrees of freedom in this configuration limit the prediction performance, based on dissimilarities between source and target domains. A detailed comparison of config 1 and 2 in shows that config 2 delivers better results until week 12 and in particular during the weeks with promotions. From this observation we can conclude that the first layer---which is un-trainable in config 2---contains general knowledge that can be transferred to the target domain. However, the restriction of the layers only makes sense up to a certain point, in our case up to week 12. After that, config 1 benefits from the higher degrees of freedom, which become more useful the more data is available for training. 
The fourth observation is that some source models have a significantly worse performance than the baseline and the other models. All points on the red line of the diagram represent outliers whose performance is more than 50\% below the model w/o\_TL. The analysis of the results shows that source 1 and source 2 are responsible for most of the outliers. In general, these two source models have by far the worst performances in all three configurations. Interestingly, these are the two products with the lowest mean hourly sales and the highest listed prices. They are listed in the same mean hourly sales cluster as the target product (\textit{"Low"}). One possible explanation for the poor performance is that the two source products have a low standard deviation of the mean hourly sales (see Table \ref{tab:product_data}).
The source models that are trained on this data can at most represent this variance of the target variable. Consequently, the transferred parameters represent only this low variance, which limits the knowledge that is transferred between the source and target domain.
Together with source 4, the two products have the lowest share of promotion data. The results show that these three source models achieved the worst results in the weeks with promotion. Since source 4 outperforms all other models in each configuration during the normal weeks, this is an indication that the share of promotion data in the source domains has an influence on the performance of the target model during the promotion weeks and that the features which represent these patterns in the source models can effectively be transferred. 
The best performance during the weeks with promotion are delivered by source 14 (config 1), source 6 (config 2.) and source 3 (config 3), all three are listed in the cluster with a \textit{"Medium"} share of promotional data.
The results show that the two products with the highest proportion of promotion data (source 5 and 8) deliver very divergent results during the promotion weeks. Thus, a high percentage of promotion data alone is not an indicator of good performance during the weeks with promotion. 

On average over all weeks, source 6 shows the best results in all three configurations. It achieves better results than the normal model in 14 and 13 weeks in config 1 and config 2 respectively. source 6 is listed in the same price cluster, but does not have exactly the same price as the target product. Overall, the source models listed in the same price cluster as the target (\textit{"Medium"} price) achieve good results. At the same time, these models are not consistently better than models in the other clusters. Thus, it can be concluded that the same price cluster is a good indicator for the similarity of the domains, but that limiting the selection criteria of the source models only to the price is insufficient. Although source 6 delivers the best result over the entire observation period, this source model does not have the lowest MSE in every single validation week and is therefore not strictly better than all other source models. From this it can be concluded that no single source model can be chosen as the optimal model. 
Another important insight can be obtained from the analysis of standard deviations of the averaged prediction results. With the exception of the source models in the cluster with \textit{"Low"} share of promotion data, the standard deviations of the TL\_models in config 1 and config 2 are significantly lower than those of the model w/o\_TL. The difference is particularly clear in the first weeks and the weeks with promotions. Thus, with TL the variance of the predictions can be significantly reduced and the models have less tendency to over-fit. 
\chapter{Discussion}\label{5}
\paragraph{RQ1} Based on the results of the case study, we can conclude the first research question: the chosen TL approach  for DNNs can be used effectively in sales forecasting. If source domains with a high similarity can be identified, the approach is a suitable method to improve the prediction performance of DNNs. The TL\_models have a) a lower variance and b) tend less to over-fit on small data sets, which addresses central challenges when applying ML models \autocite{hawkins2004problem}. The results also show that with the help of TL, complex models (config1-3) can be used despite the lack of data. Once the promotions first appear, the training data has a high bias since it is not capable of reflecting the influence of the promotions on sales. By applying TL and pre-training the models on data which already ``saw'' promotional patterns, this bias can effectively be reduced. The reduction becomes clear when comparing the MSEs of the MA baseline, the solely trained network and the TL\_models in the first week with promotion (W6). 
From the results, it can be concluded that in the context of sales forecasting ``meta'' features, such as the strong influence of promotions on sales, can be transferred between domains. 

\paragraph{RQ2} With regard to the second research question, we can conclude from the results that the similarity of the target and source tasks is, however, limited. This can be deduced from the steadily decreasing advantage of the transfer models as well as the poor performance of models in config 3. In contrast to very similar tasks, such as in the previously described applications in image classification, in our case the extensive restriction of the degrees of freedom in training leads to significantly worse results. 
The good results of config 2, especially during the first weeks and the weeks with promotions, again suggest that general knowledge is learned on the first layers of the network and can be transferred. The first layer in this configuration acts like a feature extractor and passes the extracted parameters to the trainable part of the network. This corresponds to the functionality of the first convolutional layers of a CNN \citep{JasonYosinski.}. In TL approaches with CNNs, it is common to train only on the rear layers of the neural network if the amount of available data is very small \citep{shin2016deep}. The layers containing general features of a pre-trained network are frozen during training and task-specific features of the target domain are trained on the last layers \citep{shao2014transfer}. As already described, config 1 corresponds to a pure initialization of the weights of the network at the beginning of the training. The result, that even with a large amount of available data this configuration can lead to a performance improvement, also coincides with results from other areas of TL research \citep{JasonYosinski.}. 

\begin{figure}[htbp]
 \centering
 \resizebox{\textwidth}{!}{
 \includegraphics{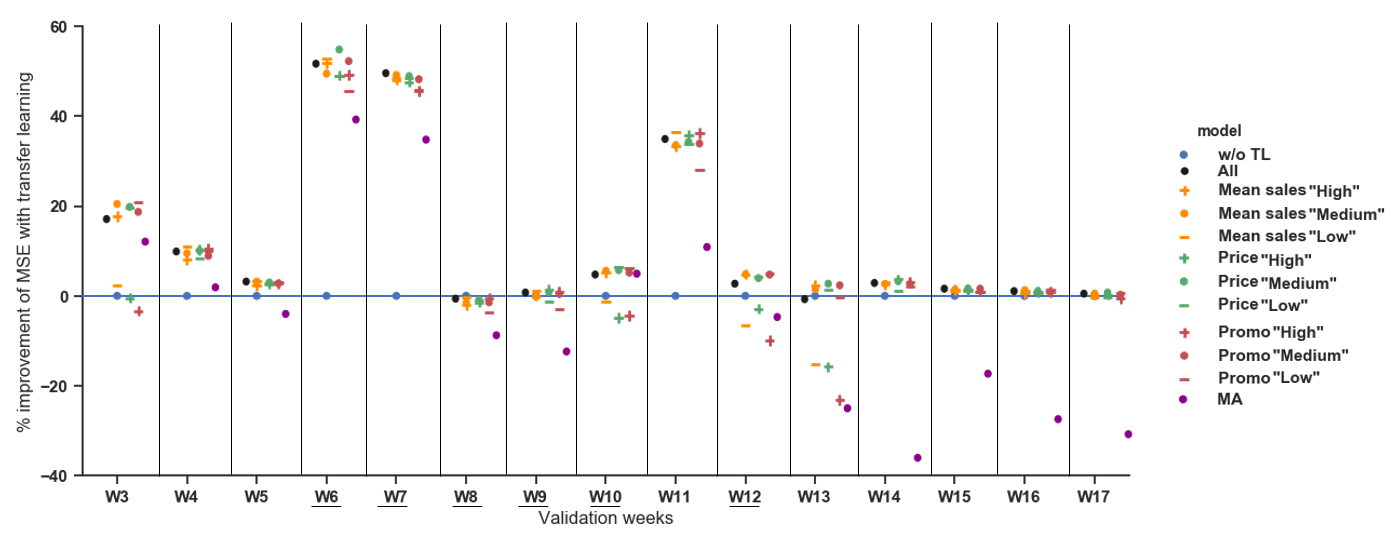}}
 \caption{Comparison of the average performances of the model w/o\_TL and the ENS\_TL\_models, after training on 30 random splits. The performance of the model w/o\_TL is used as baseline and the percentage improvement of the ENS\_TL\_models is shown} \label{fig_swarm_ens}
\end{figure}

\paragraph{RQ3} With our third research question, we wanted to investigate how the (ex-ante) known product information can be used to determine which products are best suited for the transfer. As expected, source products that have a sufficiently large share of promotion data perform better in the weeks with promotions. With regard to mean hourly sales, the results are less clear. It cannot be concluded from a high similarity of the hourly sales inevitably that the product is suitable for the transfer. This can be seen from the comparison of the performances of source 1 and source 2 with source 3 and source 4. All products have similarly low hourly sales as the target product, but deliver very different results, both during normal weeks and weeks with promotion. The poor results of source 1 and source 2 can be seen as an example for negative transfer. 
The good results of the products with a similar price (source 3, 4, 5,6 and 14) indicate that in our case study the price similarity is the most accurate indicator of task similarity. However, it must be emphasized that it is only an indicator, since the similarity dimensions examined are not disjunct. It is an intuitive assumption that products of the same product family with a similar price share similar sales patterns. It can be assumed that in expert systems the product family and the price are also the primary indicators for the selection of similar products. However, it is not possible to identify the ideal source product within this group a priori, based on similarities of price and mean hourly sales. This is illustrated by the fact that source 6 and source 14 are the two best performing models. Both are listed in the same price cluster, but source 6 has relatively low mean hourly sales, while source 14 has the highest value of all source products. Furthermore, none of the source models is outperforming all other models over the entire observation period. These findings go along with previous work examining only data similarities or even similarities of neural nets as proxy for the data. Although there might be obvious characteristics of the real-world representations of the data, influence factors for a transfer might vary depending on the individual data structures \autocite{hirt2020sequential}.

Because of this limitation, we conducted another experiment using the concept of Ensemble Learning. Ensemble Learning is a common and powerful ML approach, in which the predictions of different ML models are averaged and used for prediction \autocite{zhang2012ensemble}. 
In our case, three ensemble TL models (ENS\_TL\_models) were trained for each of the examined similarity dimensions, Mean Hourly Sales, Price and Share of Promotion Data. The ensembles are composed of the source models of the respective cluster (\textit{"Low", "Medium", "High"}) and trained on config 2. In order to verify the influence of the composition of the ensembles, one ENS\_TL\_model was trained with all listed source products. The results of this experiment are summarized in Figure \ref{fig_swarm_ens}. 
The ENS\_TL\_models show generally similar patterns as the individual source models. The performances of the ENS\_TL\_models, especially during the first weeks and the weeks with promotions, are significantly better than the model w/o\_TL and decrease over time. The evaluation of the performance of the ENS\_TL\_models provides three further interesting insights. It can be observed that the average performance of all ENS\_TL\_models, both overall and in the weeks with and without promotion, is better than that of the model w/o\_TL. 
Secondly, it can also be observed from the ENS\_TL\_models that the performance of the ensemble with a low share of promotion data in the first weeks of promotion is significantly worse than that of the other two ensembles in this cluster. The third point is that the performance of the ensemble of models in the same price cluster as the target (\textit{"Medium"}) is by far the best. The Price \textit{"Medium"} ensemble performs best compared to all models in both types of experiments, both in the weeks with and without promotions. Consequently, the ensemble strategy can be used to implement a robust TL approach that consistently delivers better results than the model w/o\_TL.

\chapter{Conclusion}
\label{6}
In this paper, we implemented a network-based Transfer Learning (TL) approach for deep neural networks (DNNs) in sales forecasting. The focus of this work was to investigate the potential of models with TL against DNNs without TL, how knowledge can be shared across source and target products and how to identify products, which are best suited for this transfer. 
In our approach, knowledge is shared by transferring layers of a pre-trained network. Within our conducted case study, this TL approach showed promising results in forecasting the sales of a newly introduced product. The TL\_models significantly improve the prediction performance in most situations with insufficient training data and during unseen promotions. By applying TL, the prediction variance can be effectively reduced and the TL\_models show an improved handling of the bias in the training data if unseen promotions occur for the first time. All three aspects are central challenges in sales forecasting and ML in general and are not limited to the context of sales forecasting in retail. 

The definition of criteria to measure similarity is essential to define the transferability between source and target domains \citep{Pan.2010}. Similarly, the proposed transfer approach can be used effectively if similar sources can be identified. The identification of subgroups, in our case based on the similarity of prices, and the investigated ensemble strategy can also be applied to other problem domains. The TL approach enables analytical knowledge sharing across domains and can be applied, when traditional forecasting techniques and ML models can not be used effectively, e.g., due to missing training data. This analytical knowledge sharing can be utilized to either support or substitute human expert systems, which are usually applied in this context. This context can be, besides the introduction of new products, the sale of seasonal products, new product categories or new branches. At the same time, the chosen approach can also be applied with other temporal granularity of the forecast or in other application fields. The application in other retail areas as well as in other contexts in which time series forecasting is used would be conceivable.

The scope of this study was limited in terms of different aspects. In regards to similarity of source and target, we did not focus on qualitative properties of the products \citep{taylor2007cross} (e.g., sweet or spicy), nor on data similarities \citep{hirt2020sequential}. We have not focused on the optimization of the hyper-parameters of the DNNs and the performance in comparison with other ML models. Therefore, one goal of our further research is to optimize the hyper-parameters of the DNN for every validation step and compare the performance with different hyper-parameter tuned ML models. 
Another limitation is that one singular model was trained for all stores. What remains to be investigated is the performance of models solely trained on the data of one store. Our approach was tested in the case study for only one product in a specific product family. Therefore, the further focus of our research is to test the approach for other newly launched products and in different product families, to verify the generalizability of our approach. A promising field of research lies ahead.
\newpage
\printbibliography

\end{document}